\documentclass[10pt,conference,a4paper]{IEEEtran}
\usepackage{amsmath,epsfig}
\usepackage{xcolor}
\usepackage{amsmath}
\usepackage{caption}

\ifCLASSINFOpdf
\else
\fi
\begin{document}

\title{Robust Visual Object Tracking with Two-Stream Residual Convolutional Networks}

\author{\IEEEauthorblockN{Ning Zhang\IEEEauthorrefmark{1}, \;
Jingen Liu\IEEEauthorrefmark{1}, \;
Ke Wang\IEEEauthorrefmark{2}, \; 
Dan Zeng\IEEEauthorrefmark{3} \; and \;
Tao Mei\IEEEauthorrefmark{4}}
\IEEEauthorblockA{\IEEEauthorrefmark{1}JD AI Research, Mountain View, USA. \quad \IEEEauthorrefmark{2}Migu Culture \& Technology, Beijing, China \quad
\IEEEauthorblockA{\IEEEauthorrefmark{3}Shanghai University, Shanghai, China \quad \IEEEauthorrefmark{4}JD AI Research, Beijing, China}}
\IEEEauthorblockA{\IEEEauthorrefmark{1}{\{ning.zhang,jingen.liu\}}@jd.com, \IEEEauthorrefmark{2}wangke@migu.cn, \IEEEauthorrefmark{3}dzeng@shu.edu.cn, \IEEEauthorrefmark{4}tmei@live.com}}

\maketitle

\begin{abstract}
The current deep learning based visual tracking approaches have been very successful by learning the target \textit{classification} and/or \textit{estimation} model from a large amount of supervised training data in offline mode. However, most of them can still fail in tracking objects due to some more challenging issues such as dense distractor objects, confusing background, motion blurs, and so on. Inspired by the human ``visual tracking'' capability which leverages motion cues to distinguish the target from the background, we propose a Two-Stream Residual Convolutional Network (TS-RCN) for visual tracking, which successfully exploits both appearance and motion features for model update. Our TS-RCN can be integrated with existing deep learning based visual trackers. To further improve the tracking performance, we adopt a ``wider'' residual network ResNeXt as its feature extraction backbone. To the best of our knowledge, TS-RCN is the first end-to-end trainable two-stream visual tracking system, which makes full use of both appearance and motion features of the target. We have extensively evaluated the TS-RCN on most widely used benchmark datasets including VOT2018, VOT2019, and GOT-10K. The experiment results have successfully demonstrated that our two-stream model can greatly outperform the appearance based tracker, and it also achieves state-of-the-art performance. The tracking system can run at up to 38.1 FPS.            

\end{abstract}


%
\IEEEpeerreviewmaketitle

\section{Introduction}



Generic visual object tracking predicts the location of a class-agnostic object at every frame of a video sequence. It is a highly challenging task due to its class-agnostic nature, background distraction, illumination discrepancy, motion blur, and many more \cite{fan2019lasot, huang2019got}. In general, a visual tracking system needs to perform two tasks simultaneously: target \textit{classification} and bounding-box \textit{estimation} \cite{danelljan2019atom}. The former task is to coarsely identify the target object region in current frame from the background, while the latter further estimates the precise bounding-box (i.e., tracker state) of the target object. 
\begin{figure}[t]
  \centering
  \centerline{\epsfig{figure=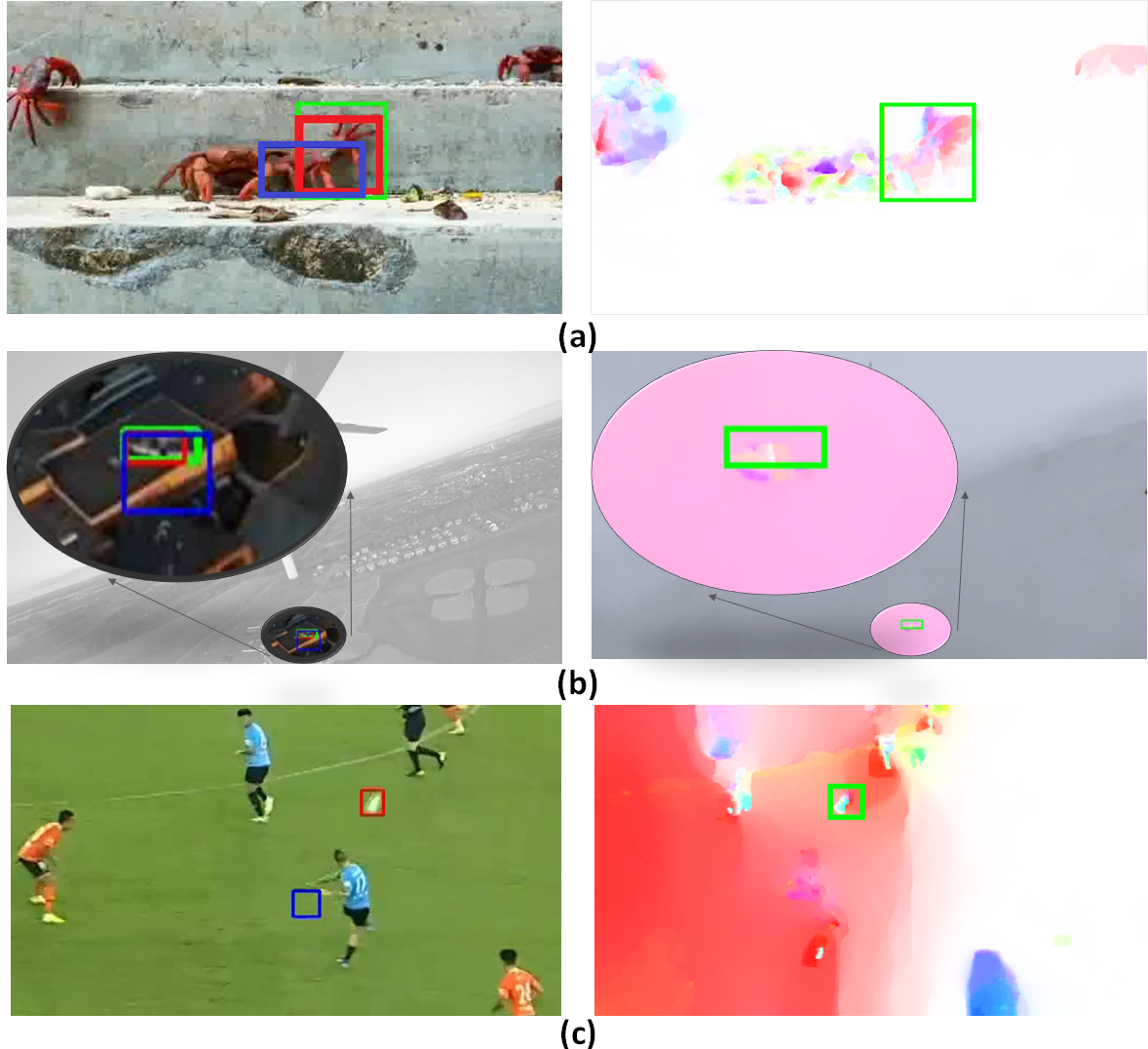,width=9cm}}
\caption{The illustration of limitations for appearance based visual tracking. The first column shows the tracking results where the \textcolor{green}{Green}, \textcolor{blue}{Blue}, and \textcolor{red}{Red} bounding box represent groundtruth, DiMP tracker, and our TS-RCN tracker, respectively. The second column illustrates the HSV-color visualization of the optical flow. In all three cases, the target optical flow has a different pattern than that of its local background. Row (a) shows dense similar objects (i.e. crabs) as distractors; Row (b) shows confusing background textures as distractors (i.e., the flying drone blends with background buildings); Row (c) shows the target (i.e., soccer ball) has motion blurs. This figure is best viewed in PDF format.}
\label{fig:ts_vs_rgb}
\end{figure}

Recently, researchers have made great progress in visual object tracking by exploiting the effective power of deep convolution networks. The Siamese tracking approaches \cite{bertinetto2016fully} leverage a large amount of supervised data to learn a more general region \textit{similarity} measurement in offline mode, which enables tracking to be performed by searching image regions most similar to the target template. Due to the lack of background appearance (e.g., distractor objects ), however, the Siamese approaches are inferior to deal with unseen objects and distractor objects. To address these limits, Bhat et al. \cite{bhat2019learning} propose a discriminative learning architecture (DiMP) which is able to fully exploit both target and background appearance.      


Although DiMP is trained to separate the background from target, it may still fail when the background becomes more confusing and challenging. As shown in Fig. \ref{fig:ts_vs_rgb} (a) and (b), DiMP is not able to track the targets (blue bounding box) due to same-class distractors (i.e., similar crabs) and confusing background texture, respectively. Additionally, the tracking can be disconnected if the current frame has motion blurs on the target. For example, the soccer ball is blurred due to high speed as shown in \ref{fig:ts_vs_rgb} (c). All the aforementioned issues can happen in most deep learning based tracking approaches where models are solely built on the target appearance. 

In contrast, humans can effectively leverage an object's motion to separate it from the background or other similar objects when its appearance is less distinguishable. To address the aforementioned issues, we propose a Two-Stream Residual Convolutional Network (TS-RCN) for visual object tracking, which is able to leverage the complementary information from both appearance and motion streams. This two-stream mechanism has been successfully applied to other video understanding tasks \cite{simonyan2014two, feichtenhofer2016convolutional}. To the best of our knowledge, however, we are the first to study an end-to-end trainable two-stream system for visual tracking. Although \cite{gladh2016deep} uses both CNN based motion and appearance features for tracking, their conventional tracking system treats deep features as a regular feature similar to hand crafted ones such as HOG. In other words, the pre-trained (on UCF101 \cite{soomro2012ucf101} and ImageNet \cite{russakovsky2015imagenet}) deep features are not trainable. As a result, the tracking system is extremely slow and less effective. 

In this work, we build our two-stream network on the DiMP tracker \cite{bhat2019learning}. It can also be integrated with other deep learning based trackers. Our TS-RCN computes optical flow for each frame and formulates it as another two-channel input in addition to the three-channel appearance. Features extracted from both streams using CNN are concatenated as one feature for tracking. Unlike \cite{bhat2019learning}, we employ the ResNeXt \cite{xie2017aggregated} as our backbone for better feature extraction, which adopts the split-transform-merge strategy to improve network performance (i.e., expanding the network from its width). 

We have extensively evaluated our TS-RCN on some widely used benchmark datasets such as VOT2018 \cite{Kristan2018a}, VOT2019 \cite{Kristan2019a}, and GOT-10K \cite{huang2019got}. All experiment results have demonstrated that two-stream visual tracking can greatly outperform appearance based tracking across all major evaluation metrics. By adopting Farneback \cite{farneback2003two} for optical flow estimation, our TS-RCN can run up to 38.1 FPS under Nvidia GTX 1080 GPU. Additionally, to demonstrate the capability of long-term tracking, we also propose a two-stream based framework to automatically initialize the lost tracker.                

In summary, we have made the following contributions:
\begin{itemize}
  \item We are the first to propose an end-to-end trainable Two-Stream Residual Convolutional Network to capture both appearance and motion cues for visual tracking. It can work with most deep learning based trackers.
  \item Unlike previous tracking approaches, our TS-RCN exploits a ``wider'' residual network ResNeXt as its feature extraction backbone to further improve the tracking performance.
  \item We propose a two-stream tracker initialization for long-term video tracking, and demonstrate this capability with soccer ball tracking.
  \item We have thoroughly evaluated our TS-RCN on most popular evaluation benchmark datasets, and achieved the state-of-the-art (STOA) performance in terms of evaluation metrics.
\end{itemize}

\section{Related Works}
\label{sec:related}

Generic visual object tracking has made great progress, thanks to the development of a variety of deep learning based approaches. The SiamFC tracker \cite{bertinetto2016fully} is the first attempt to leverage a large amount of labeled data for offline tracker training, which greatly improves the performance of conventional online trackers. Following the idea of SiamFC tracker, more recent works \cite{SiamDW_2019_CVPR, li2019siamrpn++, xu2020siamfc++, li2018high} have successfully developed to leverage various deep learning based backbone to further improve the tracking performance. For example, SiamDW \cite{SiamDW_2019_CVPR} adopts Residual Convolutional Networks \cite{he2016deep} as its backbone for feature extraction and has outperformed previous SiamFC \cite{bertinetto2016fully}.   


Due to the lack of mechanisms to explicitly train a tracker to distinguish the target from background distractors, however, the Siamese approaches generally suffer a low tracking robustness \cite{danelljan2019atom,bhat2019learning}. In other words, the Siamese approaches have a less discriminative target \textit{classifier}. To address this issue, many efforts have been made to develop discriminative online/offline target \textit{classification} and \textit{estimation} \cite{bolme2010visual,danelljan2015learning,danelljan2016beyond,danelljan2019atom,bhat2019learning} to distinguish the target from background. Inspired by correlation methods like the Discriminative Correlation Filters (DCF) \cite{bolme2010visual}, which matches region-level deep features to the target template, ATOM tracker \cite{danelljan2019atom} and DiMP tracker \cite{bhat2019learning} propose Residual Convolutional Networks (ResNet) \cite{he2016deep} based end-to-end frameworks to learn target \textit{estimation} and \textit{classification} offline, which makes visual tracking more robust. 

\begin{figure*}[t]
  \centering
  \centerline{\epsfig{figure=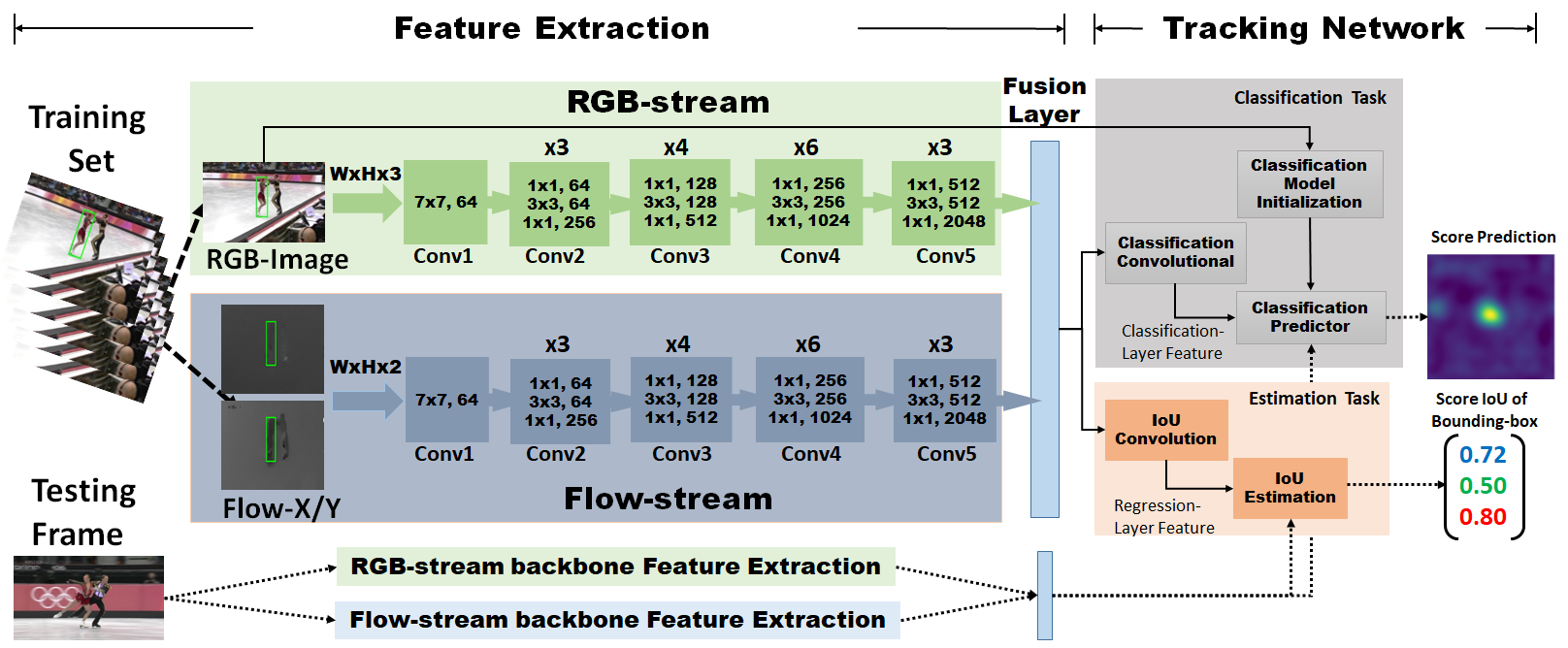,width=18 cm}}
\caption{TS-RCN Architecture with two stages: Feature Extraction and Tracking Network.}
\label{fig:TS-RCN_arch}
\end{figure*}

\begin{figure*}[t]
  \centering
  \centerline{\epsfig{figure=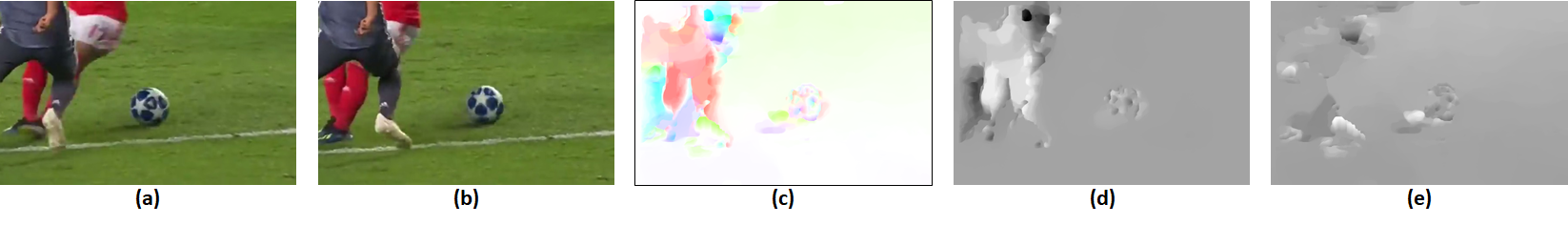,width=15cm}}
\caption{Optical flow visualization: (a-b)\: consecutive video frames of a targeted soccer ball. (c): Color visualization based on displacement vector's magnitude and direction, using the HSV color-space. (d-e): horizontal and vertical displacement vector fields $d_{u}^{t}$, and $d_{v}^{t}$, respectively, with higher intensity representing positive values.}
\label{fig:optical_flow_demo}
\end{figure*}

Although motion is an important cue for video understanding, there are only a few attempts to exploit motion cues for visual object tracking in most recent deep learning based tracking approaches. Wu \textit{et al.} \cite{wu2020motion} use Kalman-filter to model the target motion along with the Siamese trackers. Instead of integrating the kalman-filter into the end-to-end Siamese network, they use it to verify/rectify the Siamese tracker. Hence, it is less effective. A more relevant work by Gladh \textit{et al.} \cite{gladh2016deep} proposes to fuse deep motion features with deep appearance features as well as other conventional features like HOG feature into a DCF tracker. The deep motion and appearance features are pre-trained on UCF101 for action classification and ImageNet for image classification, respectively. Therefore, the deep features are not trainable, which make them no different from other conventional features. Their tracking system is not a deep learning based tracking approach, and running at an extreme slow speed ( i.e., less than 1.0 FPS). In contrast, our approach is an end-to-end trainable deep learning based visual tracking system, which can run up to 38.1 FPS with much higher tracking accuracy. 


Our TS-RCN tracking approach is inspired by some previous two-stream deep learning networks for video understand and action recognition \cite{simonyan2014two,feichtenhofer2016convolutional, wang2016temporal, carreira2017quo}. Typically, the two-streams refers to two input sources. It can be traced back to the idea of ``two-stream hypothesis'' proposed in \cite{goodale1992separate}. Such hypothesis describes the human visual cortex contains two pathways: the ventral stream preforming object recognition and the dorsal stream recognizing motions. 



\section{TS-RCN Architectures}
\label{sec:tsRCN}

In this section, the architecture of the TS-RCN is presented.
As depicted in Figure \ref{fig:TS-RCN_arch}, the end-to-end trainable network can be viewed as two stages: Feature Extraction and Tracking Network. At the Feature Extraction stage, both appearance and optical flow are inputs to two separated residual convolutional networks (i.e. RGB-stream and the optical-flow-stream), respectively. Two types of features are concatenated together via a weighted sum strategy, which forms the input to the Tracking Network for target \textit{classification} and \textit{estimation}. In our current implementation, we adopt the DiMP \cite{bhat2019learning} as our tracking network. Nevertheless, the two-stream architecture is generic and can be applied to most existing deep learning based visual tracking algorithms, such as the Siamese trackers \cite{SiamDW_2019_CVPR,li2019siamrpn++,xu2020siamfc++}.

\subsection{Optical Flow}

Optical flow is generally employed to capture the motion of objects in a video sequence. In the proposed TS-RCN architecture, the pixel-wise dense optical flows are computed as pixel displacement vectors for each frame. As shown in Figure \ref{fig:optical_flow_demo}, the flow's two fields $d_{u}^{t}$ and $d_{v}^{t}$ are calculated from two consecutive frames $t$ and $t-1$. Let $d^{t}(x,y)$ denote the displacement vector at position $(x,y)$ for frame $t$. The horizontal direction $d_{u}^{t}$ and vertical direction $d_{v}^{t}$ of the optical flows are illustrated in Figure \ref{fig:optical_flow_demo} (d) and (e), respectively.

There are various implementations of optical flow computing, such as Farneback \cite{farneback2003two}, TV-L1 \cite{perez2013tv}, FlowNet2.0 \cite{ilg2017flownet}, and so on. Since our tracking does not require high-precision optical flow estimation, we implemented a faster Farneback version (GPU based), which can run at 40 FPS at 768x576 resolution. In addition, we also tested the Total Variation (TV) based regularization with L1-norm (TV-L1) algorithm for better precision with 30 FPS at 320x240 resolution.    


During our TS-RCN training, we can leverage the pre-trained ResNet on ImageNet to the RGB-stream feature extraction. However, there is no pre-trained ResNet for optical flow features. To mimic the behavior of RGB-stream and treat two-streams equally, we discretize the optical flow into an interval from 0 to 255 using a linear transformation, which makes the range of the optical flow value the same as that of the RGB stream. This transformation unifies both RGB and optical flow streams. As a result, the pre-trained ResNeXt model with ImageNet can be applied to both streams.  

\subsection{Two-Stream Architecture}

After the optical flow computation and preprocessing, both RGB and optical flow streams take their respective input and feed them to the backbone networks. In our experiment, the RGB-stream takes RGB color-channels as input, while the flow-stream takes the optical flow u/v channels as input. Each of the RGB-stream or flow-stream can be a ResNet or its variations such as ResNeXt or Wide residual networks (WRNs) \cite{zagoruyko2016wide}. In Figure \ref{fig:TS-RCN_arch}, we use ResNet blocks for illustration. A fusion layer is applied to combine features of the two streams. We adopted a weighted sum mechanism which was introduced in action recognition. \cite{simonyan2014two, wang2016temporal}



The two-stream combined feature is input to the \textit{classification} and \textit{estimation} tasks. At the \textit{classification} branch, the feature goes through a convolutional block to extract the classification-layer feature. It is then used to train the classification predictor. In addition, an RGB-only model initialization is used when the initial frame $F^{(0)}$ with precise region of interest (ROI) is given, the ROI pooling operation is conducted to get the same size feature as the classification-layer feature for the predictor model. This initialization effectively reduces the optimization recursion for the classification prediction. Simultaneously, in the bounding-box \textit{estimation} branch, a different IoU convolutional block takes the two-stream fused feature to extract the bounding-box regression-layer feature. This newly extracted feature is fed into the IoU network \cite{jiang2018acquisition} based \textit{estimation} model. Since it is an end-to-end trainable mechanism, the loss of combined regression from \textit{estimation} and the \textit{classification} are back-propagated through the two-stream structure. 

Formally, given a video \textbf{V}, we randomly select two segments \{ $M_{mod}, M_{train}$ \}. $M_{mod}$ data segment is selected in a fashion that is always prior to $M_{train}$ along the time course. Each set $M = \{I_j, F_j, B_j\}_{j=1}^{N_{frames}}$, consists of images $I_j$, optical flow images $F_j$, and their paired corresponding target bounding-boxes $B_j$ at the current frame $j$. The optical flow of each $F_j$ is calculated from the paired frames $(I_{j-1}, I_j)$.

\{ $S_{mod}, S_{train}$ \} are the modulate samples and training samples. Our setup follows the DiMP tracker \cite{bhat2019learning} in that $S_{mod}$ is used to provide a model predictor as a preprocessing to predict the discriminative feature and maintain the generalization for the future unseen $S_{train}$ samples.

$S_{train}$ is formulated below, where $x_{j}= x_{ts}(I_j, F_j)$ is the two-stream feature extraction of the backbone residual networks, and $c_j$ is the center coordinate of the bounding-box $B_j$.

\begin{equation}\label{eqn:s_train}
S_{train} = \{(x_{ts}(I_j, F_j), c_j) : (I_j, F_j, B_j) \in M_{train}\}
\end{equation}

The combined two-stream feature $x_{j}$ is defined as the weighted sum of $R_{rgb}(I_{j})$ and $R_{flow}(F_{j})$, where $R_{rgb}(\cdot)$ and $R_{flow}(\cdot)$ are the residual convolutional network for the RGB and flow streams respectively.
\begin{equation}\label{eqn:ts_weighted_avg}
x_{j} = \lambda_{rgb} * R_{rgb}(I_{j}) + \lambda_{flow} * R_{flow}(F_{j})
\end{equation}

The obtained two-stream feature $x_{j}$ are used to feed the \textit{classification} and \textit{estimation} networks. The total loss function $L_{tot} = \beta L_{cls} + L_{est}$, where $L_{cls}$ is the \textit{classification} loss, and $L_{est}$ is the bounding-box \textit{estimation} loss. The target \textit{classification} is based on a hinge based regression error, given a confidence score $s$ and the target region $z$. The $T$ is the threshold.

\begin{equation}\label{eqn:hinge_error}
l(s,z)=
\begin{cases} 
      s-z & z > T \\
      max(0, s) & z\leq T 
   \end{cases}
\end{equation}

The above hinge error is used in calculating the $L_{cls}$.The confidence score $s$ is represented as the input feature at the $i_{th}$ iteration's target model $f^{(i)}$, where $x$ is the two-stream feature, and the $*$ denotes the convolution.

\begin{equation}\label{eqn:L_cls}
L_{cls}=\frac{1}{N_{iter}} 
\sum_{i=0}^{N_{iter}} \sum_{(x, c) \in S_{test}}\bigg\| l\big(x*f^{(i)}, z_c \big) \bigg\|^{2}
\end{equation}

The \textit{estimation} loss $L_{est}$ minimizes a prediction error of the following equation using mean-squared error function $g(*)$, following the IoU-Net \cite{jiang2018acquisition}. The function $c(*)$ is the modulation function generating the modulation vector.  $x_0$ and $B_0$ are extracted two-stream features and the corresponding bounding-box from the first frame in $M_{mod}$.  The function $IoU(*)$ is the IoU model prediction using the IoU-Net, with the input of extracted two-stream feature $x$ and its corresponding bounding-box $B$ from $M_{train}$

\begin{equation}\label{eqn:L_est}
L_{est}(B) = g(c(x_0, B_0), IoU(x, B))
\end{equation}

\subsection{Tracking}
At the inference stage, the tracking sequence's current frame is input to both RGB-stream and optical-flow-stream for backbone feature extraction. Fusion layer gets the expected value of the two-stream input with a weighted average the same as the training stage. The calculated two-stream feature is used in classification predictor to generate a score prediction on the target object's location. The feature is also used in the IoU prediction where proposals were calculated and ranked with the IoU model for the best bounding-box estimation. 

\begin{figure}[t]
  \centering
  \centerline{\epsfig{figure=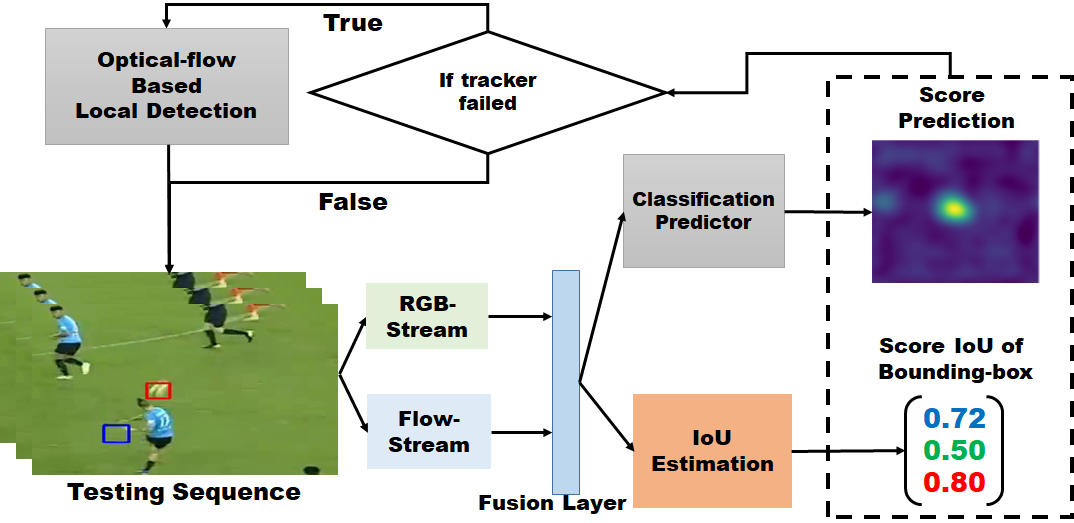,width=9 cm}}
\caption{Tracker re-initialization via an optical flow local search object detection. }
\label{fig:inference_reinit}
\end{figure}

\subsection{Long-video Tracking} 
In long-video SOT tasks, tracker initialization with object detection also suffers from the similar appearance-only limitation due to the target object blurring, speed variations, similar-appearance distractors and etc. In addition to the tracker model, we furthermore propose an optical flow based initialization mechanism which takes the feedback from the tracking results and adaptively initialize the tracker with optical flow based local detection.  Figure \ref{fig:inference_reinit} demonstrates this re-initialization. When the target \textit{classification} and \textit{estimation} collectively triggers a failed tracker, an optical flow based re-initialization is implemented to refine the object's bounding-box. For instance, the prediction score or the IoU score is lower than a threshold. As an illustration, the testing images from Figure \ref{fig:inference_reinit} depicts this situation. The red-color bounding-box indicates the before and after failed tracker re-initialization which retrieved the soccer ball successfully. The blue-color bounding-box indicates a failure in tracking the target without such re-initialization. Video examples of these will be provided as the supplementary material.



\section{Experiment Results}
\label{sec:exp}
To verify the effectiveness of our proposed TS-RCN for visual object tracking, we conducted extensive experiments on some broadly used SOT benchmark datasets, such as VOT2018, VOT2019, and GOT-10K. Our approach was implemented in Python with pyTorch, and all experiments were run on Nvidia GTX 1080. Our models were mainly trained on three datasets, namely GOT-10K \cite{huang2019got}, LaSOT \cite{fan2019lasot}, and ImageNetVid \cite{russakovsky2015imagenet}. The trackingNet and Microsoft COCO image dataset,\footnote{Still image only dataset is not suitable for extracting optical flows.} are not used. The ImageNet pre-trained models were used for both appearance and motion stream. With GPU-enabled Farneback optical flow computation, TS-RCN tracker can run at real-time mode on VOT2018 benchmark. Specifically, for the best performing backbone ResNeXt-50, its tracking speed is 21.74 FPS. Furthermore, TS-RCN's speed can achieve up to 38.1 FPS with ResNet-18 backbone. 





\subsection{Evaluation Metrics}

In our experiments, we used tracking \textit{accuracy} and \textit{robustness} to measure performance. Both metrics were adopted by VOT-2013 Challenge and have become widely used since then \cite{kristan2013visual}. 

Tracking \textit{accuracy} is simply defined as the overlap between the tracker's bounding-box and ground-truth bounding-box. Given a continuous tracker \textit{i} (no lost frames), the per-frame \textit{accuracy} at frame \textit{t} is computed as $\phi_{t}(i) = \frac{A_{t}^{G} \cap A_{t}^{T}}{A_{t}^{G} \cup A_{t}^{T}}$, where $A_{t}^{T}$ and $A_{t}^{G}$ are the estimated and the ground-truth bounding-box, respectively. Due to the stochastic nature of a tracker, a tracking system will be repeatedly evaluated $N_{rep}$ times on one dataset, and the actual per-frame \textit{accuracy} $\Phi_{t}(i)$ is the average of $\phi_{t}(i)$ over $N_{rep}$ times. Now, let $M_{valid}$ be the total number of frames for tracker $\textbf{i}$, the tracking \textit{accuracy} ($\rho_{A}(i)$) of a tracker \textit{i} is the average of $M_{valid}$ per-frame \textit{accuracy} $\Phi_{t}(i)$.

Tracking \textit{robustness} measures how reliable of a tracker without losing the target. It is linked to the tracking failures, which is the total number of tracking lost (i.e., the tracker drifts away from the target). Let $F(i,k)$ be the failure times of the \textit{i-th} tracker in experiment \textit{k}, then the \textit{robustness} $\rho_{R}(i)$ of tracker \textit{i} is the average over all $N_{rep}$ repeated experiments. The overall \textit{accuracy} and \textit{robustness} across the entire testing sequence can be calculated as the weighted average of the per-sequence performance with weights proportional to the lengths of the sequences.  



In VOT2015 Challenge \cite{kristan2015visual}, the Expected Average Overlap (\textit{EAO}) was designed as a principled combination of \textit{accuracy} and \textit{robustness}. Since then, it has been adopted as one of the main evaluation criteria for tracking performance. More details can be found in \cite{kristan2016novel}. In the following experiments, we use VOT-toolkit\footnote{https://github.com/votchallenge/vot-toolkit} and pysot-toolkit\footnote{https://github.com/STVIR/pysot} which are widely used to calculate the \textit{EAO}, \textit{accuracy} and \textit{robustness}.



\subsection{Ablation Study}  
Unless otherwise specified, we employed ResNeXt-50 backbone for feature extraction in our experiments, and trained TS-RCN tracker on dataset LaSOT and GOT-10K. All trackers are evaluated on VOT2018 testing dataset.   

\textbf{Two-Stream Hyper Parameter.} As we can see from Equation \ref{eqn:ts_weighted_avg}, two hyper parameters $\lambda_{rgb}$ and $\lambda_{flow}$ are used to balance the effect of two streams in the tracking results. When $\lambda_{rgb} = 1$ and $\lambda_{flow} = 0$, TS-RCN becomes a DiMP tracker, which serves as our baseline approach. We conducted a grid-search-like approach to find the best value combination of $\lambda_{rgb}$ and $\lambda_{flow}$. Table \ref{tb:tsResNeXt50} shows the performance of TS-RCN in terms of \textit{EAO}, \textit{accuracy}, \textit{robustness}. When $[\lambda_{rgb}, \lambda_{flow}]=[0.5,0.5]$, the system achieves best performance in all metrics. Additionally, we notice that two-stream systems generally outperform two single-stream systems ($\lambda_{flow}=0$ or $\lambda_{rgb}=0$) across most evaluation metrics. It clearly illustrates that the TS-RCN tracker can work better than the single-stream tracker.

\begin{table}
\centering
\resizebox{\columnwidth}{!}{%
\begin{tabular}{lccccccccc}
  \hline
 \textbf{[$\lambda_{rgb}$, $\lambda_{flow}$]} & \textbf{[1, 0]} & \textbf{[3/4, 1/4]} & \textbf{[2/3, 1/3]} & \textbf{[1/2, 1/2]} & \textbf{[1/3, 2/3]} & \textbf{[1/4, 3/4]} & \textbf{[0, 1]}\\ 
  \hline
EAO $\uparrow$ & 0.379 & 0.370 & 0.354  & \textbf{\textcolor{red}{0.459}} & \textbf{\textcolor{blue}{0.398}} & 0.367 & 0.031 \\ 
Accuracy $\uparrow$ & \textbf{\textcolor{blue}{0.557}} & 0.515 & 0.511 & \textbf{\textcolor{red}{0.579}} & 0.520 & 0.552 & 0.366 \\ 
Robustness $\downarrow$ &  0.187 & 0.182 & 0.178 & \textbf{\textcolor{red}{0.139}} & \textbf{\textcolor{blue}{0.148}} & 0.201 & 3.467\\ 
   \hline
\end{tabular}%
}
\caption{The results of TS-RCN trackers with various weighting combinations of $[\lambda_{rgb}, \lambda_{flow}]$. The settings of $[1,0]$ and $[0, 1]$ are actually two single-stream modes, i.e., the RGB and optical flow trackers, respectively. The top-2 performance results are colored in  \textbf{\textcolor{red}{Red}} and \textbf{\textcolor{blue}{Blue}}, respectively.}
\label{tb:tsResNeXt50}
\end{table}

\begin{figure}[t]
  \centering
  \centerline{\epsfig{figure=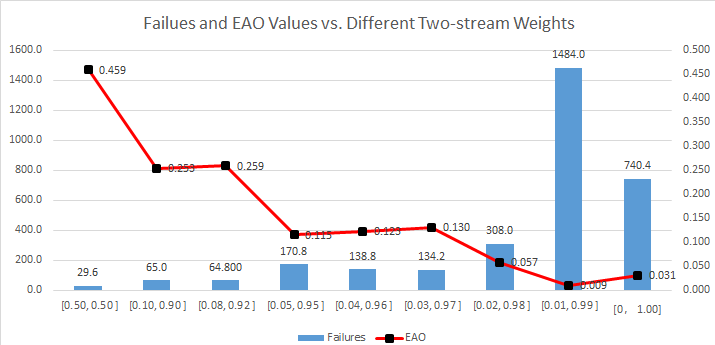,width=9cm}}
\caption{ \textit{EAO} and Failures for different $[\lambda_{rgb}, \lambda_{flow}]$.}
\label{fig:flow_impact}
\end{figure}

It is also interesting to look at the plummeted performance when only single-stream optical flow is employed for tracking. To further analyze this situation, we performed another group of experiments which look into the fine-scale changes of $\lambda$ values. Figure \ref{fig:flow_impact} depicts the tracking performance with $\lambda$ changing at a scale of 0.01, from $[0.05, 0.95]$ to $[0, 1.0]$. The bar value shows the failure numbers for each $\lambda$ configure, and the red-line with black-dot shows the corresponding \textit{EAO} value. When $[\lambda_{rgb}, \lambda_{flow}]$ changes to $[0.02, 0.98]$, Failure rate significant increases and \textit{EAO} decays drastically. This observation shows that appearance is a primary stream for TS-RCN tracker given current experimental configurations. We conjecture that it may be due to the pre-train process of flow-stream network. We adopted the pre-trained RGB model for the flow-stream. To solve this issue, we can separately train a pre-trained optical flow model, which we will leave to our future work.    



\begin{table}
\centering
\resizebox{\columnwidth}{!}{%
\begin{tabular}{rcccc}
  \hline
 \textbf{ResNet Depth} & \textbf{18} & \textbf{50} & \textbf{101} & \textbf{152} \\ 
  \hline
EAO $\uparrow$ & 0.345 & \textbf{\textcolor{red}{0.419}} & \textbf{\textcolor{blue}{0.383}} & 0.233 \\ 
\hline
Params (millions) & 11.69 & 25.56 & 44.56 & 60.19 \\
\hline
\end{tabular}%
}
\caption{ResNet depth impact of the performance with \textit{EAO} scores and corresponding networks' number of parameters (in millions). Weights $[\lambda_{rgb}, \lambda_{flow}] = [0.5, 0.5]$ is used.}
\label{tb:ablation_depth}
\end{table}


\begin{table}
\centering
\resizebox{\columnwidth}{!}{%
\begin{tabular}{rccc}
  \hline
 \textbf{DBs} & \textbf{GOT-10k} & \textbf{GOT-10k + LaSOT} & \textbf{GOT-10k + LaSOT + ImgNetVid} \\ 
  \hline
\large{EAO $\uparrow$} & \textbf{\textcolor{blue}{\large{0.383}}} & \large{\textbf{\textcolor{red}{0.459}}} & \large{0.378}\\
\hline
\end{tabular}%
}
\caption{EAO scores of dataset choices on the proposed TS-RCN with $[\lambda_{rgb}, \lambda_{flow}] = [0.5, 0.5]$.}
\label{tb:ablation_dataset}
\end{table}


\begin{table*}
\centering
\resizebox{\textwidth}{!}{%
\begin{tabular}{lccccccccccccccc}
  \hline
&\textbf{LADCF} & \textbf{MFT} & \textbf{SiamRPN} & \textbf{DRT} &\textbf{RCO} & \textbf{UPDT} &\textbf{ECO} &\textbf{SiamFC} & \textbf{ATOM} & \textbf{SiamFC++} & \textbf{DaSiamRPN} & \textbf{SiamMask} & \textbf{SiamRPN++} & \textbf{DIMP-50} & \textbf{TS-RCN}\\ 
&\cite{xu2019learning} & \cite{bai2018mft} & \cite{li2018high} & \cite{sun2018correlation} & \cite{Kristan2018a} & \cite{bhat2018unveiling} & \cite{DanelljanCVPR2017} & \cite{bertinetto2016fully} & \cite{danelljan2019atom} & \cite{xu2020siamfc++} & \cite{zhu2018distractor} & \cite{wang2019fast} & \cite{li2019siamrpn++} & \cite{bhat2019learning} & ours\\ 
  \hline
EAO $\uparrow$ & 0.389 & 0.385 & 0.383 & 0.356 & 0.376 & 0.378 & 0.280 & 0.187 & 0.401 & 0.426 & 0.326 & 0.387 & 0.414 & \textbf{\textcolor{blue}{0.422}} & \textbf{\textcolor{red}{0.459}}\\ 
Accuracy $\uparrow$ & 0.505 & 0.508 & 0.587 & 0.519 & 0.507 & 0.536 & 0.487 & 0.505 & 0.590 & 0.587 & 0.569 & \textbf{\textcolor{red}{0.642}} & 0.600 & \textbf{\textcolor{blue}{0.602}} & 0.579\\ 
Robustness $\downarrow$ & 0.159 & \textbf{\textcolor{blue}{0.140}} & 0.276 & 0.201 & 0.155 & 0.184 & 0.276 & 0.585 & 0.204 & 0.183 & 0.337 & 0.295 & 0.234 & 0.162 & \textbf{\textcolor{red}{0.139}}  \\ 
   \hline
\end{tabular}%
}
\caption{Comparison with the STOA on VOT2018.}
\label{tb:stoa_vot2018}
\end{table*}

\begin{table}
\centering
\resizebox{\columnwidth}{!}{%
\begin{tabular}{lcccccccc}
  \hline
&\textbf{DRNet} & \textbf{Trackyou} & \textbf{ATP} & \textbf{SiamRPN++} &\textbf{SiamMask} & \textbf{SiamDW\_ST} & \textbf{DIMP-50} & \textbf{TS-RCN}\\ 
&\cite{Kristan2019a} & \cite{Kristan2019a} & \cite{Kristan2019a} &  \cite{li2019siamrpn++} & \cite{wang2019fast} & \cite{SiamDW_2019_CVPR} & \cite{bhat2019learning} & ours\\ 
\hline
EAO $\uparrow$ & \textbf{\textcolor{blue}{0.393}} & \textbf{\textcolor{red}{0.394}} &  \textbf{\textcolor{blue}{0.393}} & 0.282 & 0.287 & 0.297 & 0.342 & 0.375\\ 
Accuracy $\uparrow$ & 0.602 &  \textbf{\textcolor{blue}{0.610}} & \textbf{\textcolor{red}{0.649}} & 0.598 & 0.596 & 0.597 & 0.600 & 0.582\\ 
Robustness $\downarrow$ & \textbf{\textcolor{red}{0.261}} & 0.268 & 0.291 & 0.482 & 0.461 & 0.467 & 0.321 &  \textbf{\textcolor{blue}{0.262}} \\ 
\hline
\end{tabular}%
}
\caption{Comparison with the STOA on VOT2019.}
\label{tb:stoa_vot2019}
\end{table}

\textbf{Backbone Depth.} Network depth is another important factor that affects the performance of the TS-RCN. In this group of experiments, we examined how the tracking performance changes with various backbone depths. As shown in Table \ref{tb:ablation_depth}, ResNet achieves better performance when depth is 50, although deeper network generally produces better results. Deeper structure means many more parameters, which needs a lot more training data to avoid over-fitting. Given our current training data, depth of 101 or of 152 may cause over-fitting.

\textbf{Backbone Architecture.}
As shown in the above experiments, the residual convolutional networks obtained better tracking accuracy at depth of 50. In addition to the ResNet, various residual convolutional networks have been developed, such as the ``width'' of the network. In this ablation study, we compare the tracking performance of ResNet, ResNeXt, and WRNs at depth of 50. The classic ResNet-50 has width of bottleneck 4. The ResNext-50 has additional hyper parameter cardinality 32. The WRNs-50 has additional widening factor 2. We first verified that for all structures, two-stream approach achieved the best performance with $[\lambda_{rgb}, \lambda_{flow}] = [0.5, 0.5]$. For the ResNet-50 backbone, we achieved 0.419 \textit{EAO}, 0.571 \textit{accuracy}, and 0.168 \textit{robustness}.  For the WRNs-50 backbone, we obtained 0.390 \textit{EAO}, 0.568 \textit{accuracy}, and 0.195 \textit{robustness}. Both structures show inferior performance than the counterpart ResNeXt-50 in Table \ref{tb:tsResNeXt50}. This ablation study demonstrates that increasing the cardinality with the split-transform-merge strategy improves the network performance.


\textbf{Training Dataset.} Deeper network may work better with more training data. But the quality or type of the dataset may also matter. We conducted a set of experiments, in which we used various combination of datasets for training. Table \ref{tb:ablation_dataset} presents tracking performance comparison in terms of \textit{EAO}. GOT-10K is currently the largest annotated video datasets with 563 categories, from 10,000 video segments with more than 1.5 million manually-labeled bounding-box \cite{huang2019got}. LaSOT is a dataset with 70 categories, from 1400 videos. \cite{fan2019lasot}. ImageNetVid is part of the ImageNet ILSVRC2015 competition, consisting of a 30-category objects data from 4500 videos, with a total of more than one million annotated frames \cite{russakovsky2015imagenet}. As we can see, the combination of LaSOT and GOT-10K achieves the best performance with \textit{EAO} score 0.459. It is worth noting that it is not always the case that more datasets, the better performance. As indicated in Table \ref{tb:ablation_dataset} that when all three datasets used, the \textit{EAO} score dropped to 0.378. One possible cause is that the training is object-appearance dependent. Hence, more category with balanced data will help to improve the performance. Category increase with less balanced data may jeopardize the performance. ImageNetVid has only 30 categories which has a large overlap with the previous two datasets. Hence, the performance tends to plateau and even decrease in our experiment.  



\textbf{Two-Stream Fusion Strategy.} We have two schemes to fuse the RGB and flow streams. The features are extracted separately by two identical but independent networks from two streams. It can be called ``late-fusion''. Alternatively, we can stack 3 RGB channels with 2 optical flow channels, and then feed it into one common ResNeXt for feature extraction. This scheme is so-called ``early-fusion''. In terms of \textit{EAO}, late-fusion greatly outperforms early-fusion (0.459 vs 0.356). This could be because the optical flow and RGB are different representations in nature, one for pixel movement, and one for RGB values. Simply stacking them with a single backbone model diminishes each stream's value and hence affects the overall performance.



\subsection{State-of-the-Art Comparison} 



\textbf{VOT2018 and VOT2019 datasets.} Following previous SOT evaluation, we evaluated our best tracking model on both VOT2018 and VOT2019 datasets, which each containing 60 challenging testing sequences.     


As shown in Table \ref{tb:stoa_vot2018}, our TS-RCN tracker with ResNeXt-50 achieves the best tracking results on VOT2018 in \textit{EAO} and \textit{robustness} metrics, as compared to some recent popular trackers such as SiamMask, DiMP, ATOM, and so on. It clearly illustrates that motion cues can greatly improve tracking robustness, which is critical for many practical applications. Relatively speaking, \textit{accuracy} is less important since it only measures the degree of intersection with the ground truth. More specifically, our approach improves the \textit{EAO} of DiMP-50 ( single-stream version of our TS-RCN ) by 3.7\%, and reduces the failure (\textit{robustness}) by 2.3\%, although our model was trained with less training data than \cite{bhat2019learning}. For a fair comparison, we re-train DiMP tracker with the same training data as ours, and it achieves 0.385 \textit{EAO}, 0.563 \textit{accuracy} and 0.209 \textit{robustness}, which are all worse than that of ours. 

There are fewer results on VOT2019 dataset. Per Table \ref{tb:stoa_vot2019}, our approach achieves the top result in terms of \textit{robustness}. Our results are comparable to the top performers on VOT2019 benchmark, and better than that of recently developed STOA tracking approaches. 


\textbf{GOT-10k dataset \cite{huang2019got}}. On GOT-10K dataset, researchers use average overlap (\textit{AO}) and success rate (\textit{SR}) to evaluate all trackers. \textit{AO} denotes the average of overlaps between the estimated bounding-boxes and ground-truth. \textit{SR} measures the percentage of successfully tracked frames where the overlaps exceed a threshold (e.g. 0.5, 0.75). We used the GOT-10k training split for both network training and validation, and the test split for testing. The test split has 180 videos. There is no overlap of object classes between the train and the test, which prevents over-fitting of an individual class. Hence, the purpose of this evaluation focuses on testing the generalization capabilities of trackers on unseen object classes. To ensure a fair comparison, the presented trackers are trained and validated using the same subset drawn from the GOT-10k training set. This strategy prevents using external datasets or different training set split to ensure the difference is only caused by the employed backbone structures. 

Table \ref{tb:stoa_got10k} presents the results. For the average overlap, baseline DiMP tracker achieves 59.7\% and two-stream ResNet-50 achieves 60.8\% and a relative gain of 1.1\%. ResNeXt-50 achieves the highest with 60.7\% and a relative gain of 1.0\%. It also tops the success rate @0.5 and @0.75 overlap threshold among all models. This experiment using the same setting on different datasets verifies the generalization ability of the proposed TS-RCN.



\begin{table}
\centering
\resizebox{\columnwidth}{!}{%
\begin{tabular}{rcccc}
  \hline
& \textbf{DIMP-50} &  \textbf{TS-RCN} & \textbf{TS-RCN} & \textbf{TS-RCN} \\ 
& \cite{bhat2019learning} & \textbf{ResNet-50}  & \textbf{ResNeXt-50} & \textbf{WRNs-50} \\ 
  \hline
SR\_0.75 (\%) & 45.8 &  \textbf{\textcolor{blue}{46.0}} &  \textbf{\textcolor{red}{47.7}} & 35.2\\ 
SR\_0.50 (\%) & 71.1 &  \textbf{\textcolor{red}{71.6}} &  \textbf{\textcolor{red}{71.6}} & 61.1\\ 
AO (\%)  & 59.7 &  \textbf{\textcolor{red}{60.8}} &  \textbf{\textcolor{blue}{60.7}} & 52.2 \\ 
   \hline
\end{tabular}%
}
\caption{GOT10k testing result in terms of average overlap (AO), success rates (SR) at overlap thresholds 0.5 and 0.75.}
\label{tb:stoa_got10k}
\end{table}

\subsection{Visualization}
\begin{figure*}[hbt!]
  \centering
  \centerline{\epsfig{figure=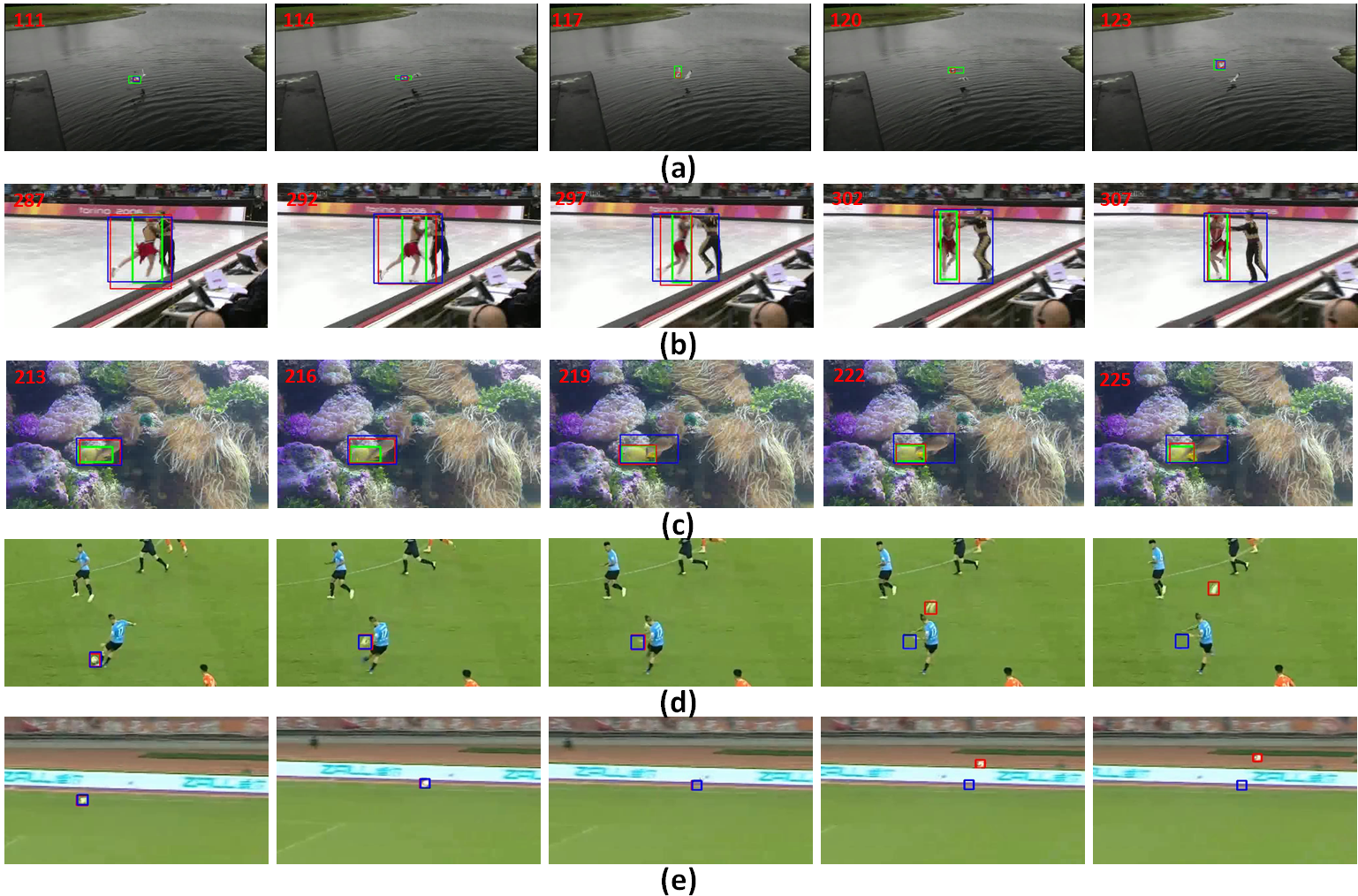,width=17.8 cm}}
\caption{Visualization of the TS-RCN tracker in \textbf{\textcolor{red}{Red}} and the baseline DiMP tracker in \textbf{\textcolor{blue}{Blue}}. The groundtruth is in  \textbf{\textcolor{green}{Green}}. Row (a-c): each sequence skips certain fixed-length frames rather than showing every consecutive frame in-between. For reference, the frame number is given on the top left corner. Row (d-e): long-video tracking with optical flow based local detection.}
\label{fig:exp_visualization}
\end{figure*}

Some visualization examples are given in Figure \ref{fig:exp_visualization}. The top three rows are sequences from the VOT2018 benchmark. Row (a) demonstrates the blurred target object in appearance. Row (b-c) show the distractors with same category appearance. Although not the same visual appearance, RGB single stream tends to confuse the distractor from the true target if they are the same category and in close proximity.



For the long-term video tracking, we focus on soccer ball tracking in real matches. The soccer ball may be blurred and deformed due to the high speed, or blended with background distractors. The bottom two rows of Figure \ref{fig:exp_visualization} are examples of optical flow based local detection, which retrieves the soccer ball using the motion information. In comparison, RGB single-stream approach fails in tracking. Row (d) demonstrates the occluded soccer ball by the player. The soccer ball is also blurred and deformed. Row (e) illustrates the distractor (advertisement board) which has the same appearance. We will also provide the corresponding long-video tracking clips in the supplementary materials.


\section{Conclusion}
\label{sec:con}
We propose a Two-Stream Residual Convolutional Network which strategize to combine the RGB appearance and the optical flow motion inputs. Our proposed architecture is based on the ResNeXt residual networks as the backbone structure and can be integrated with existing trackers using end-to-end training. This strategy has been tested on VOT2018, VOT2019, and GOT-10k benchmarks, where it outperformed the baselines that only use RGB single-stream, as well as other STOAs provided by the benchmarks. 





\bibliographystyle{IEEEtran}
\bibliography{ms.bib}




\end{document}